\def\year{2022}\relax
\documentclass[letterpaper]{article} % DO NOT CHANGE THIS
\usepackage{aaai22}  % DO NOT CHANGE THIS
\usepackage{times}  % DO NOT CHANGE THIS
\usepackage{helvet}  % DO NOT CHANGE THIS
\usepackage{courier}  % DO NOT CHANGE THIS
\usepackage[hyphens]{url}  % DO NOT CHANGE THIS
\usepackage{graphicx} % DO NOT CHANGE THIS
\usepackage{natbib}  % DO NOT CHANGE THIS AND DO NOT ADD ANY OPTIONS TO IT
\usepackage{caption} % DO NOT CHANGE THIS AND DO NOT ADD ANY OPTIONS TO IT
\DeclareCaptionStyle{ruled}{labelfont=normalfont,labelsep=colon,strut=off} % DO NOT CHANGE THIS
\usepackage{algorithm}
\usepackage{subfig}
\usepackage{algpseudocode}
\usepackage{commath}
\usepackage{array,multirow}
\usepackage{amsmath}
\usepackage{amsfonts}
\usepackage{amssymb}
\usepackage{comment}

\usepackage{stmaryrd}
\usepackage{bbm}

\usepackage{newfloat}
\usepackage{listings}
\floatstyle{ruled}
\newfloat{listing}{tb}{lst}{}
\floatname{listing}{Listing}
\title{Graph-wise Common Latent Factor Extraction for Unsupervised Graph Representation Learning}
\title{Graph-wise Common Latent Factor Extraction for Unsupervised Graph Representation Learning}
\author {
    Thilini Cooray,
    Ngai-Man Cheung
}
\affiliations {
    Singapore University of Technology and Design (SUTD)\\
   
    thilini\_cooray@mymail.sutd.edu.sg, ngaiman\_cheung@sutd.edu.sg
}

\begin{document}

\maketitle

%another title, global property inspired common latent factor extraction for graph representation learning

\begin{abstract}
Unsupervised graph-level representation learning plays a crucial role in a variety of tasks such as molecular property prediction and community analysis, especially when data annotation is expensive. Currently, most of the best-performing graph embedding methods are based on Infomax principle. The performance of these methods highly depends on the selection of negative samples and hurt the performance, if the samples were not carefully selected. Inter-graph similarity-based methods also suffer if the selected set of graphs for similarity matching is low in quality. To address this, we focus only on utilizing the current input graph for embedding learning. We are motivated by an observation from real-world graph generation processes where the graphs are formed based on one or more global factors which are common to all elements of the graph (e.g., topic of a discussion thread, solubility level of a molecule). We hypothesize extracting these common factors could be highly beneficial. Hence, this work proposes a new principle for unsupervised graph representation learning: Graph-wise Common latent
Factor EXtraction (GCFX). We further propose a deep model for GCFX, deepGCFX, based on the idea of reversing the above-mentioned graph generation process which could explicitly extract common latent factors from an input graph and achieve improved results on downstream tasks to the current state-of-the-art. Through extensive experiments and analysis, we demonstrate that, while extracting common latent factors is beneficial for graph-level tasks to alleviate distractions caused by local variations of individual nodes or local neighbourhoods, it also benefits node-level tasks by enabling long-range node dependencies, especially for disassortative graphs.\footnote{Our source code:  https://github.com/thilinicooray/deepGCFX} 

\end{abstract}

\section{Introduction}

Graph structured data has been very useful in representing a variety of data types including social networks \cite{PhysRevE.69.026113}, protein-protein interactions \cite{ppifirst}, scene graphs \cite{krishnavisualgenome}, customer purchasing patterns \cite{Bhatia16} and many more \cite{7814302,Cooray_2020_CVPR,DBLP:journals/sigpro/LiuC21}. In this work, we focus on graph-level representation learning. It is crucial for tasks like molecular property identification \cite{NIPS2015_5954} and community classification  \cite{DBLP:conf/kdd/YanardagV15}, and they are useful for applications such as drug discovery, material design and recommendation systems. Availability of task-specific labels plays a significant role in graph representation learning. However, annotations are very expensive \cite{yang2019analyzing,WIEDER2020,DBLP:conf/iclr/SunHV020} due to many specialized fields in which graphs are utilized (e.g., biological sciences, quantum mechanics). Therefore, unsupervised graph representation learning has become crucial. 

Unsupervised graph level representation learning has a very rich literature consisting of several main directions. Skip-gram influenced graph embedding methods (node2vec \cite{DBLP:conf/kdd/GroverL16}, sub2vec \cite{DBLP:conf/pakdd/AdhikariZRP18}, graph2vec \cite{DBLP:journals/corr/NarayananCVCLJ17}) only rely on neighbourhood information and lose the advantage of using node features, making them less effective. Kernel methods (Random Walk (RW) \cite{DBLP:conf/colt/GartnerFW03}, Shortest Path (SP) \cite{1565664}, Graphlet Kernel (GK) \cite{DBLP:journals/jmlr/ShervashidzeVPMB09}, DDGK \cite{DBLP:conf/www/Al-RfouPZ19}, GCKN \cite{DBLP:journals/corr/abs-2003-05189}) and graph proximity methods (UGraphEmb \cite{DBLP:conf/ijcai/BaiDQMG0SW19}) use pair-wise inter-graph similarity calculations making them more effective, but less efficient. Quality of learnt embeddings by this method heavily relies on the quality and variety of other graphs with which it compares. Contrastive learning (InfoGraph \cite{DBLP:conf/iclr/SunHV020}, CMV (Multi-view) \cite{icml2020_1971}, GCC (Extra data) \cite{qiu2020gcc} and GraphCL(Augmentations) \cite{DBLP:conf/nips/YouCSCWS20}) is the newest addition which is based on the Infomax principle \cite{DBLP:journals/computer/Linsker88} which aims at obtaining an output which has maximum mutual information with the input. The main drawback of these methods \cite{DBLP:conf/nips/GrillSATRBDPGAP20} is that their heavy reliance on the selection procedure of negative samples for model performance. A careful selection of task-wise negative samples is required to obtain good performance. While both inter-graph similarity and contrastive methods achieve state-of-the-art for graph embedding learning, both of them suffer very high if the quality of other graphs they compare with is low. 

Autoencoder \cite{Baldi1989NeuralNA, DBLP:conf/nips/HintonZ93} based embedding methods solve this weakness by only utilizing its current input for representation learning. However, existing graph autoencoder models \cite{DBLP:journals/corr/KipfW16a,DBLP:conf/ijcai/PanHLJYZ18,DBLP:conf/iccv/ParkLCLC19} are only aimed at node-level modelling. {\bf In particular, these methods are unable to extract graph-wise common latent factors.}  Other issues of these methods are their overemphasis on proximity information \cite{icml2020_1971} and the inability to differentiate features  \cite{DBLP:conf/eccv/TianKI20}. Feature differentiation is very important when using embeddings in downstream tasks because equally treated features could add noise and redundancy which leads to performance degradation.

These weaknesses of existing work motivate us to research an approach that could both differentiate features crucial for graph-level representations as well as capable of learning embeddings by only utilizing the current input sample. Although GVAE \cite{DBLP:journals/corr/KipfW16a} is inadequate to fulfil graph-level feature differentiation, it empowers single sample-based learning. Hence, we are motivated to follow a generative-based mechanism while addressing the specific requirements to obtain a discriminative graph representation which GVAE lacks on.  

{\bf Graph-wise common latent factor extraction (GCFX) Motivation.} To draw inspiration for our generative-based approach, we observe two real-world graph formation examples. An online discussion thread can be represented as a graph where nodes represent users who have participated in the discussion thread, and edges represent interactions among users in the thread \cite{DBLP:conf/kdd/YanardagV15}. This graph initializes with a single user who wants to discuss a particular topic and grows with nodes when subsequent users start responding about this topic. For the second example, a chemical compound can be represented as a graph where the nodes are atoms and edges are chemical bonds. Inverse molecule design \cite{Sanchez-Lengeling360, kuhn1996inverse, zunger2018inverse} is a molecule generation process that is initiated with the desired properties to be included in a molecule such as solubility and toxicity levels. De novo \cite{schneider2013novo, doi:10.1021/acs.jcim.8b00839} inverse molecule design method initiates with the desired ranges of those properties and iteratively adds atoms and chemical bonds conditioned on those properties to form molecular graphs.

The key observation from these examples is that each node and edge added to the graph was conditioned on one or more common graph factors. The topic is the common global factor for all elements of the discussion thread and toxicity and solubility levels are common for the entire chemical compound. We can see that although each node has its own specific information such as personal details of a user or properties of an atom, common factors cause the differentiation of one graph from another. Hence, useful for tasks like community detection and molecule selection for drug discovery.

{\bf Graph-wise common latent factor extraction.} Motivated by this, we hypothesize extracting these common factors could be highly beneficial for a discriminative graph representation. Hence, this work proposes graph-wise common factor extraction in a latent manner. We further propose deepGCFX: a novel autoencoder-based architecture that can explicitly extract common latent factors from the entire graph incorporating feature differentiation to autoencoders. Our enhanced decoding mechanism which enforces utilizing common latent factors for graph reconstruction, also regularizes normal autoencoder's heavy dependence on proximity. 

\noindent We summarize our contributions as follows:
\begin{itemize}
    \item We propose GCFX: a novel principle for unsupervised graph representation learning based on the notion of graph-wise common latent factors inspired by real-world examples. 
    \item Existing autoencoder models are unable to learn graph-wise common factors due to their inability of feature differentiation. Therefore, we propose deepGCFX: a novel autoencoder-based approach with iterative query-based reasoning and feature masking capability to extract common latent factors.
    \item We empirically demonstrate the effectiveness of deepGCFX in extracting graph-wise common latent factors. 
    \item To the best of our knowledge, this is the first graph embedding learning method based on GCFX. We show that deepGCFX can achieve state-of-the-art results in {\em unsupervised graph-level representation learning} as shown in standard downstream tasks.
    \item By extracting common factors from non-common, deepGCFX enables long-distance inter-node information sharing ability for node representation learning achieving best results for unsupervised node representation learning on disassortative graphs.
\end{itemize}

\section{Methodology}

\subsection{Graph generation process } \label{sec:graphgen}

Let $\mathbb{D} = \{\mathbb{G}, C_f, L_f\}$ be the set that consists of graphs and their ground truth common and uncommon generative factors. We would call uncommon factors as local, hence the notation $L$. Each graph $G = (V,A)$, contains a set of nodes $V$ and $A$ is the adjacency matrix. $C_f$ and $L_f$ represent two sets of generative factors: $C_f$ contains common factors $\mathbf{c_f} \subset C_f$ common for the entire graph (e.g., discussion topic) and $\mathbf{l_f} \subset L_f$ represents local factors which can differ from local patch to patch (e.g., user information). In this work we assume, $\mathbf{c_f}$ and $\mathbf{l_f}$ are conditionally independent given $G$, where $p( \mathbf{c_f,l_f} | G) = p(\mathbf{c_f}|G) \cdot p(\mathbf{l_f}|G)$. We assume that the graph $G$ is generated using a true world generator which uses the ground truth generative factors: $p(G|\mathbf{c_f,l_f}) = \textit{Gen}(\mathbf{c_f,l_f})$.

\subsection{GCFX: Graph-wise common latent factor extraction}

\begin{figure*}[h]
	\centering
	\includegraphics[scale=0.54]{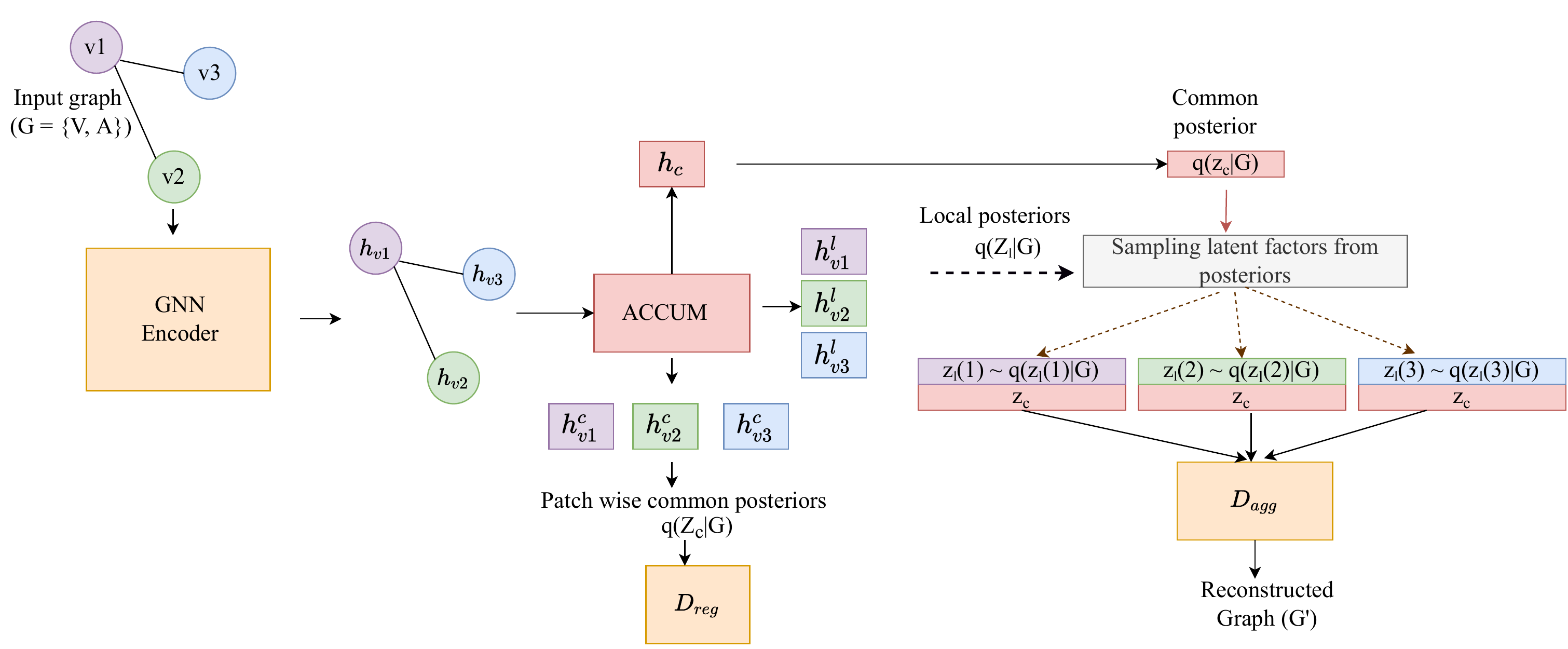}
	\caption{deepGCFX architecture: An input graph $G$ is first sent through a GNN Encoder to obtain individual node representations aggregated with neighbours and then the ACCUM procedure is conducted to filter graph-wise common ($\mathbf{h}^{c}$) and local factors ($\mathbf{h}^{l}$) from each patch to obtain a single graph-wise common latent factor representation $\mathbf{h}_c$ (ACCUM is described Fig. \ref{fig:accum}). Sampled local latent factors from their respective posteriors $\mathbf{z}_l(j) \sim q(\mathbf{z}_l(j)|G), ~ \forall j \in \{1 \dots |V|\}$ are combined with the common latent $\mathbf{z}_c \sim q_(\mathbf{z}_c|G)$ and input to the decoder $\mathcal{D}_{agg}$ to reconstruct $G$. $\mathcal{D}_{reg}$ is used to enforce $\mathbf{z}_c$ to contain $G$ related factors. deepGCFX is optimized using the loss function in Eq.\ref{eq:finalloss}.}
	\label{fig:gcfx}
\end{figure*}

We focus on the novel problem of graph-wise common factor extraction. Although we only focus on extracting common latent factors, identifying local factors is essential to filter them out. Hence, our goal is to develop an unsupervised deep graph generative model which can learn the joint distribution of graph $G$, the set of generative factors $\mathbf{Z}$, using only the samples from $\mathbb{G}$. This should be learnt in a way that the set of latent generative factors can generate the observed graph $G$, such that $p(G|\mathbf{Z}) \approx p(G|\mathbf{c_f,l_f}) = \textit{Gen}(\mathbf{c_f,l_f})$. A suitable approach to fulfil this objective is to maximize the marginal log-likelihood for the observed graph $G$ over the whole distribution of latent factors $\mathbf{Z}$.

\begin{equation}
 \max_{\theta} \mathbb{E}_{p_{\theta}(\mathbf{Z})} [\log
p_{\theta}(G|\mathbf{Z})] 
\label{eq:likelihood}
\end{equation}

For an observed graph $G$, the inferred posterior probability distribution of the latent factors $\mathbf{Z}$ can be described as $q_\phi( \mathbf{Z}|G)$. However, the graph generation process we described above assumes two independent sets of generative factors representing common and local information of a graph, from which we are explicitly interested in common factors. Therefore, we consider a model where the latent factor set $\mathbf{Z}$ can be divided into two independent latent factor sets as $\mathbf{Z} = (\mathbf{Z}_c, \mathbf{Z}_l)$. $\mathbf{Z}_c$ represents the latent factors that capture the graph-wise common generative factors of $G$ and $\mathbf{Z}_l$ captures its local counterpart. Therefore, we can rewrite our inferred posterior distribution as follows:
\begin{equation}
\begin{aligned}
q_\phi( \mathbf{Z}|G) & = q_\phi( \mathbf{Z}_c, \mathbf{Z}_l|G) = q_\phi( \mathbf{Z}_c|G)q_\phi(\mathbf{Z}_l|G)
\label{eq:ourposterior1}
\end{aligned}
\end{equation}

We discuss these two posteriors in detail:
$q_\phi( \mathbf{Z}_c|G)$ and $q_\phi(\mathbf{Z}_l|G)$.
The graph $G$ consists of $|V|$ number of nodes. In a graph data structure, each node is not isolated. They are connected with their neighbours and propagate information. Therefore, we use the term {\em patch} to indicate the local neighbourhood centered at each node which the node interacts with. 
Therefore,  $q_\phi( \mathbf{Z}_c|G)$ and $q_\phi( \mathbf{Z}_l|G)$ are the posterior distributions of all these $|V|$ patches. However, common latent posterior is common for all $|V|$ patches, as the graph $G$ was originally generated with $\mathbf{c_f}$ common for all $V$. Hence, we propose to use a single latent $\mathbf{z}_c$ to capture the common generative factors.  In particular, we use $q_\phi( \mathbf{z}_c|G)$ to model this single posterior. On the other hand, the factors which contribute to generate each patch can vary significantly. Therefore, in this model, we assume the local latent factors are independent. 
Therefore, we update Eq. \ref{eq:ourposterior1} as: 

\begin{equation}
\begin{aligned}
q_\phi( \mathbf{Z}|G) & = q_\phi(\mathbf{z}_c,\mathbf{Z}_l|G)  = q_\phi( \mathbf{z}_c|G) \prod_{i=1}^{|V|} q_{\phi}(\mathbf{z}_l(i)|G)
\label{eq:ourposterior2}
\end{aligned}
\end{equation}

Here, $\mathbf{z}_l(i)$ are the latent factors that capture the local generative factors for a patch centered at node $i$.
Now, our objective is to make sure the latent factors sampled from common and local latent posterior distributions  can capture the common and local generative factors $\mathbf{c}_f$ and $\mathbf{l}_f$ separately.

\subsection{A constrained optimization formulation for GCFX}

Now we try to match common and local generative factors $\mathbf{c}_f$ and $\mathbf{l}_f$ to their respective priors $p(\mathbf{z}_c)$ and $p(\mathbf{z}_l)$ separately. We select unit Gaussians ($\mathcal{N}(0,1)$) as priors. Based on our modeling of common and local factors in Eq. \ref{eq:ourposterior2}, we can rewrite Eq.\ref{eq:likelihood} as a constrained optimization as follows \cite{DBLP:conf/iclr/HigginsMPBGBML17}:

\begin{equation}
\begin{aligned}
 \max_{\theta,\phi}\quad & \mathbb{E}_{G\sim \mathbb{G}}\Big[  \mathbb{E}_{q_\phi(\mathbf{z}_c,\mathbf{Z}_l|G)}  [\log
p_{\theta}(G|\mathbf{z}_c,\mathbf{Z}_l)] \Big] \\
\textrm{s.t.} \quad & KL(q_{\phi}(\mathbf{z}_c|G)\parallel p(\mathbf{z}_c)) < \epsilon \\
& KL(q_{\phi}(\mathbf{Z}_l|G)\parallel p(\mathbf{Z}_l)) < \eta \\
\label{eq:consopt}
\end{aligned}
\end{equation}

\noindent where $\epsilon$ and $\eta$ are strengths of each constraint. 
Following \citet{DBLP:conf/iclr/HigginsMPBGBML17}, 
Eq.\ref{eq:consopt} can be written to obtain the variational evidence lower bound (ELBO) of a Graph Variational Autoencoder (GVAE) \citep{DBLP:journals/corr/KipfW16a} (Here we use GVAE because of our input is a graph) as follows with $\beta$ and $\gamma$ coefficients:

\begin{equation}
\begin{aligned}
\mathcal{F}(\theta,\phi;G,\mathbf{z}_c,\mathbf{Z}_l,\beta,\gamma) 
& \geq \mathcal{L}(\theta,\phi;G,\mathbf{z}_c,\mathbf{Z}_l,\beta,\gamma) \\
& = \mathbb{E}_{q_\phi(\mathbf{z}_c,\mathbf{Z}_l|G)}  [\log
p_{\theta}(G|\mathbf{z}_c,\mathbf{Z}_l)]  \\
& - \beta ~ KL(q_{\phi}(\mathbf{z}_c|G)\parallel p(\mathbf{z}_c)) \\
& - \gamma ~ KL(q_{\phi}(\mathbf{Z}_l|G)\parallel p(\mathbf{Z}_l))
\label{eq:firstelbo}
\end{aligned}
\end{equation}

Based on Eq.\ref{eq:ourposterior2}, we can expand the KL divergence term $KL(q_{\phi}(\mathbf{Z}_l|G)\parallel p(\mathbf{Z}_l))$ and rewrite our objective function for a single graph $G$ as:

\begin{equation}
\begin{aligned}
 \mathcal{L}(\theta,\phi;G,\mathbf{z}_c,\mathbf{Z}_l,\beta,\gamma) 
& = \mathbb{E}_{q_\phi(\mathbf{z}_c,\mathbf{Z}_l|G)}  [\log
p_{\theta}(G|\mathbf{z}_c,\mathbf{Z}_l)]  \\
& - \beta ~ KL(q_{\phi}(\mathbf{z}_c|G)|p(\mathbf{z}_c)) \\
& - \gamma ~ \sum_{i=1}^{|{V}|}KL(q_{\phi}(\mathbf{z}_l(i)|G) \parallel p(\mathbf{z}_l(i)))
\label{eq:finalobj}
\end{aligned}
\end{equation}

Overall, {\bf the learning objective of GCFX is to maximize this lower bound} for all the graphs in a minibatch $\mathbb{G}_b$ from the full dataset $\mathbb{G}$:
\begin{equation}
{\mathcal L}_{\theta,\phi}(\mathbb{G}_b ) = \frac{1}{|\mathbb{G}_b|}\sum_{r=1}^{|\mathbb{G}_b|}  \mathcal{L}(\theta,\phi;G_r,\mathbf{z}_c,\mathbf{Z}_l,\beta,\gamma)
\label{eq:fullELBO}
\end{equation}

\section{deepGCFX: An autoencoder based approach for GCFX }\label{sec:archi}

Existing autoencoder models including GVAE cannot learn graph-wise common factors due to their inability to differentiate factors based on importance. Therefore, we propose deepGCFX, a novel GVAE architecture based on the GCFX principle which can extract graph-wise common factors. We propose an iterative query-based mechanism with feature masking to achieve this ability. Fig. \ref{fig:gcfx} depicts the proposed deep Graph-wise Common Factor Extractor (deepGCFX) model. We utilize $N$-layer Graph neural Network(GNN)\cite{DBLP:conf/iclr/KipfW17} as the encoder. $n^{th}$ layer of a GNN can be defined in general as 

\begin{equation}
    \mathbf{a}_v^{(n)}   = \text{AGGREGATE}^{(n)} \left( \left\lbrace \mathbf{h}_u^{(n-1)}  : u \in \mathcal{N}(v) \right\rbrace \right)
    \label{eq:gnn_layer}
\end{equation}

\begin{equation}
    \mathbf{h}_v^{(n)}   = \text{COMBINE}^{(n)} \left( \mathbf{h}_v^{(n-1)},\mathbf{a}_v^{(n)} \right)
    \label{eq:gnn_layer}
\end{equation}

\noindent where $\mathbf{h}_v^{(n)}$ is the feature vector of a patch centered at node $v \in V$ at the $n^{th}$ layer after propagating information from its neighbours $ u \in \mathcal{N}(v)$, where $(v,u) \in A$. $\mathbf{h}_v^{(0)}$ is often initialized with node features. We use the term \text{GNN} to indicate any network which uses layers described in Eq. \ref{eq:gnn_layer}. The neighbourhood aggregation function $\text{AGGREGATE}$ and node update function $\text{COMBINE}$ differs for each specific GNN architecture \cite{DBLP:conf/iclr/KipfW17, DBLP:conf/iclr/VelickovicCCRLB18, DBLP:conf/iclr/XuHLJ19}.

\subsection{ACCUM : Iterative query-based reasoning with feature masking for common latent accumulation}\label{sec:accum}

\begin{figure}[h]
	\centering
	\includegraphics[scale=0.45]{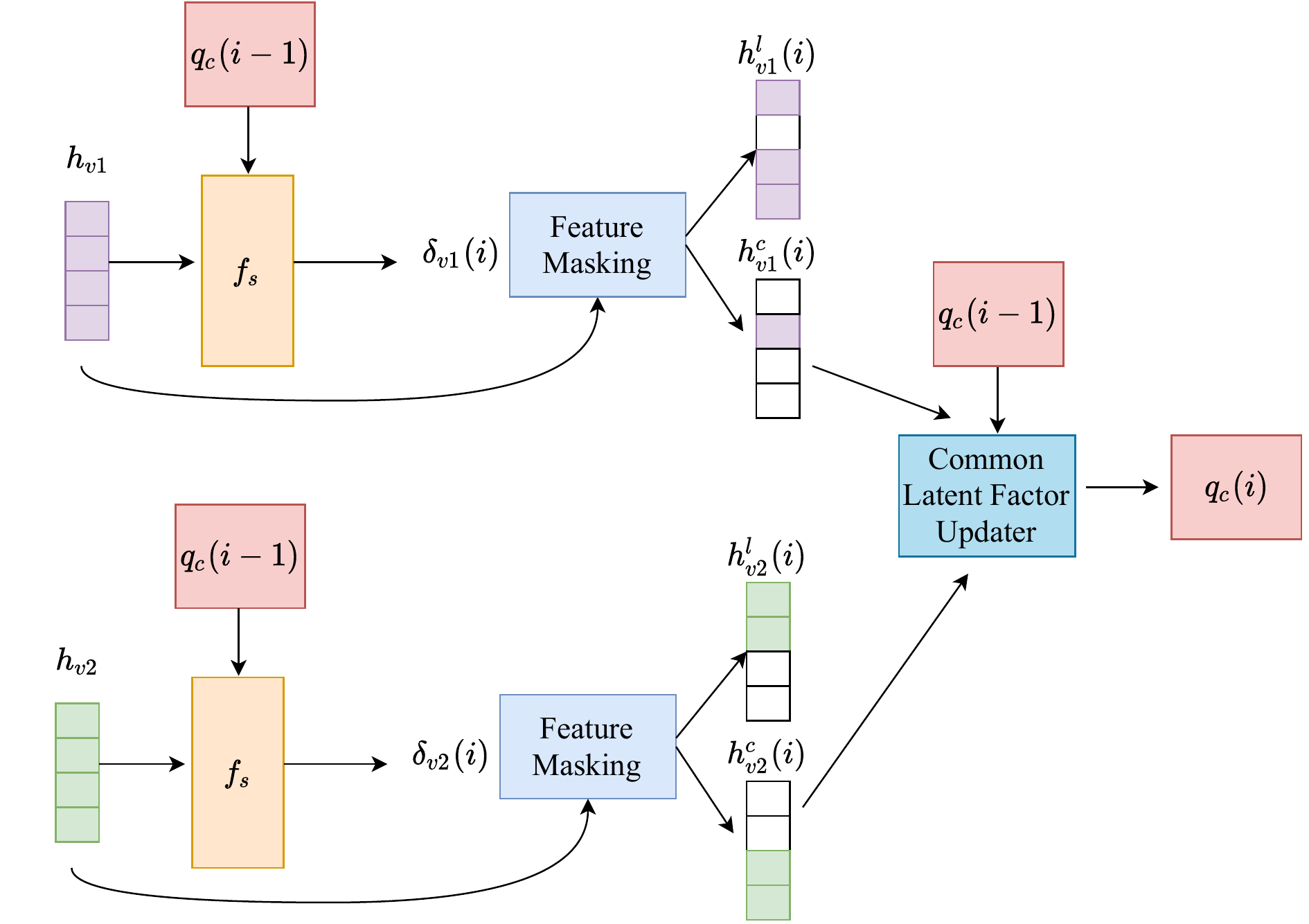}
	\caption{ ACCUM : Our main algorithmic invention to enable extraction of high-quality common latent factors for deepGCFX is based on query-based reasoning with feature masking. For this two node graph, at iteration $i$, the single common factor embedding from previous iteration used as the query $\mathbf{q}_c(i-1)$. For each node, $\delta(i)$ is calculated (Eq. \ref{eq:node_wise_weight1}) to determine its factor-wise similarity to current common latent factors. Each node's latent factors are divided into common ($\mathbf{h}^{c}$) and local ($\mathbf{h}^{l}$) next, using a $\delta(i)$ based mask (Eq. \ref{eq:node_wise_filter1}-\ref{eq:node_wise_filter4}). Node-wise $\mathbf{h}^{c}$ are accumulated using Eq.\ref{eq:c_update} to update graph-wise common latent factors for current iteration.}
	\label{fig:accum}
\end{figure}

{\bf Our main algorithmic invention is a novel mechanism to extract high-quality common latent factors based on ideas of iterative query-based reasoning and feature masking.} As discussed, GVAE cannot extract common latent factors as it cannot differentiate feature importance. To ensure the latent factors sampled from common and local posterior distributions can capture common and local factors separately, we first need a mechanism to differentiate and filter out what are common features from the output node/patch representations of the GNN encoder. To achieve this, we propose a novel mechanism that iteratively learns graph-wise common factors and extracts them from each patch and accumulates them to generate a single common factor embedding for each graph. 

We model the common factor extraction as an iterative query-based reasoning problem, where our query is the accumulated common factor representation of the graph. We use that query on our input patch representations $\mathbf{h}_v^{(n)}$ (superscript will be omitted from here onwards.) to determine which factors from $\mathbf{h}_v$ are similar to the existing common factors and filter them out from non-common factors to update the accumulated common factor representation. At each iteration $i$, the process starts with the query $\mathbf{q_c}(i-1)$ containing graph-wise common factors extracted from all the patches at iteration $i-1$.  It is used to query all patch representations $\mathbf{h}_v, v \in V$ at current iteration $i$ to identify the amount of similarity each factor of $\mathbf{h}_v, v \in V$ has with current graph-wise common factors $\mathbf{q_c}(i-1)$ which is also our query. The factor-wise similarity score for current iteration $i$, $\delta_v(i) \in \mathbb{R}^{d\_hidden}$ is calculated as:

\begin{equation}
    \delta_v(i) = \sigma(f_s([\mathbf{h}_v\mathbf{W}_k,
    \mathbf{q_c(i-1)}\mathbf{W}_q])) ,
    \label{eq:node_wise_weight1}
\end{equation}

\noindent where $\mathbf{W}_k, \mathbf{W}_q, \in \mathbb{R}^{d\_hidden \times d\_hidden}$ are projection parameters for query and the keys, $f_s$ is a non-linear network and $[\cdot]$ denotes concatenation. Then we create two masks; mask $\mathbf{m}_v^{c}(i)$ is to filter out factors of patch $v$, which are similar to current common factors $\mathbf{q_c}(i-1)$ and mask $\mathbf{m}_v^{l}(i)$ is to filter out remaining local factors. 

\begin{align}
    \mathbf{m}_v^{c}(i) &= \mathbbm{1}\llbracket \sigma(\mathbf{h}_v\mathbf{W}_k) \geq \delta_v(i) \rrbracket  ,
    \label{eq:node_wise_filter1}\\
    \mathbf{m}_v^{l}(i) &= \mathbbm{1}\llbracket \sigma(\mathbf{h}_v\mathbf{W}_k) < \delta_v(i) \rrbracket,
    \label{eq:node_wise_filter2}\\
    \mathbf{h}_v^{c}(i) &= \mathbf{m}_v^{c}(i) \odot \mathbf{h}_v\mathbf{W}_v,
    \label{eq:node_wise_filter3}\\
    \mathbf{h}_v^{l}(i) &= \mathbf{m}_v^{l}(i) \odot \mathbf{h}_v\mathbf{W}_v
    \label{eq:node_wise_filter4}
\end{align}

\noindent where $\mathbf{W}_v \in \mathbb{R}^{d\_hidden \times d\_hidden}$ is projection parameters and $\mathbf{m}_v^{c}(i), \mathbf{m}_v^{l}(i) \in \mathbb{R}^{d\_hidden} $. $\odot$ denotes element-wise multiplication. Now $\mathbf{h}_v^{c}(i), \forall v\in V$ are accumulated and used to update $\mathbf{q_c}(i-1)$ with newly identified common factors for iteration $i$ using a Gated Recurrent Unit(GRU):

\begin{align}
    \mathbf{q}_{update}(i) &= \sum_{v \in V} \mathbf{h}_v^{c}(i),\\
    \mathbf{q_c}(i) &= \text{GRU}(\mathbf{q}_{update}(i), \mathbf{q_c}(i-1))
    \label{eq:c_update}
\end{align}

This accumulation approach is depicted in Fig. \ref{fig:accum}. Once ACCUM is over in $M$ iterations, we use $\mathbf{h}_c~( = \mathbf{q_c}(M))$ and $\mathbf{h}_v^{l}(M)$ to generate parameters for our posterior distributions $q_\phi(\mathbf{z}_c|G)$ and $q_\phi(\mathbf{z}_l(j)|G), ~ \forall j \in \{1 \dots |V|\}$.

\subsection{Aggregation and Regularization based decoding}

We propose a novel decoder different from the local proximity emphasized GVAE decoder to enforce the two qualities our common latent factor $\mathbf{z}_c$ should possess; $\mathbf{z}_c$ must be common for all patches and it should be relevant to input graph $G$. We utilize two decoders to fulfil these requirements. 

\paragraph{Aggregation decoder $\mathcal{D}_{agg}$ to enforce commonality of $\mathbf{z}_c$ for all patches $v$.} 

To reconstruct the original graph properly, the model requires both common and local factors. Therefore, common and local latent factors are sampled from their respective posterior distributions ($\mathbf{z}_c \sim q_\phi(\mathbf{z}_c|G)$ and $\mathbf{z}_l(j) \sim q_\phi(\mathbf{z}_l(j)|G), ~ \forall j \in \{1 \dots |V|\}$) and sent through the decoder $\mathcal{D}_{agg}$ for reconstructing $G$. Note that the graph-wise common latent factors $\mathbf{z}_c$ is only sampled once for the entire graph using $q_\phi(\mathbf{z}_c|G)$. The aggregated reconstruction of $G$ is achieved via adjacency matrix reconstruction $A$, where $p_\theta(A_{jk}=1|\mathbf{z}_c, \mathbf{z}_l(j),\mathbf{z}_l(k )) =   \mathcal{D}_{agg}([\mathbf{z}_c,\mathbf{z}_l(j)])^{T}~\mathcal{D}_{agg}([\mathbf{z}_c,\mathbf{z}_l(k)])$.

Although $\mathcal{D}_{agg}$ enforces that $\mathbf{z}_c$ should contain factors common to all patches to enable proper graph reconstruction, $\mathcal{D}_{agg}$ cannot enforce what type of common factors $\mathbf{z}_c$ should possess. Since $\mathbf{z}_c$ is a constant for all patches, if the type of information that $\mathbf{z}_c$ should have was not enforced, there is a possibility that $\mathbf{z}_c$ gets ignored by $\mathcal{D}_{agg}$. To mitigate this, GCFX employs a {\bf Regularization decoder $\mathcal{D}_{reg}$} which enforces $\mathbf{z}_c$ that it must contain structural information about the graph $G$ by trying to reconstruct $G$ on its own as shown in Fig. \ref{fig:gcfx}.

\subsection{Training and Inference details} 
We modify the objective function in Eq. \ref{eq:finalobj} obtained in the general framework for common latent factor extraction to following $\mathcal{L}_{deepGCFX}$ for training deepGCFX in an end-to-end manner.

\begin{equation}
\begin{aligned}
\mathcal{L}_{deepGCFX} &= \mathcal{L}_{\mathcal{D}_{agg}} + \beta\mathcal{L}_{c\_prior} + \gamma\mathcal{L}_{l\_prior} + \mathcal{L}_{\mathcal{D}_{reg}}\\
& =\mathbb{E}_{q_\phi(\mathbf{z}_c,\mathbf{Z}_l|G)}  [\log
p_\theta(G|\mathbf{z}_c,\mathbf{Z}_c)]\\
& - \beta ~ KL(q_\phi(\mathbf{z}_c|G)\parallel p(\mathbf{z}_c)) \\
& - \gamma ~ \sum_{j=1}^{|{V}|}KL(q_\phi(\mathbf{z}_l(j)|G) \parallel p(\mathbf{z}_l(j))) \\
& + \mathbb{E}_{q_\phi(\mathbf{z}_c,\mathbf{Z}_l|G)}  [\log p_\theta(G|\mathbf{Z}_c)]
\label{eq:finalloss}
\end{aligned}
\end{equation}

\paragraph{Inference.} After training deepGCFX as above, we utilize $\mathbf{z}_c \sim q_\phi(\mathbf{z}_c|G)$ as learnt single common latent factor representation for graph $G$ and $\mathbf{z}_l(j) \sim q_\phi(\mathbf{z}_l(j)|G), ~ \forall j \in \{1 \dots |V|\}$ as local non-common node/patch-specific latent factor representations for downstream tasks.

\begin{table*}[h]
		\caption{Mean 10-fold cross validation accuracy on graph classification. Results in \textbf{bold} indicate the best accuracy for both inter-graph similarity based and non-inter-graph similarity based methods separately. \underline{Underlined} results show the second best performances.We follow strictly the experiment and evaluation setup and datasets as in \cite{DBLP:conf/iclr/SunHV020, icml2020_1971} for deepGCFX and GVAE baseline. Results of other methods are taken from their papers.}
		\label{table:graph_eval}
		
		\begin{center}
			\begin{small}
				
					\begin{tabular}{lcccccc}
						\hline
						\textbf{\textsc{dataset}} & \textbf{\textsc{mutag}} & \textbf{\textsc{ptc-mr}} & \textbf{\textsc{imdb-bin}} &\textbf{ \textsc{imdb-mul}} & \textbf{\textsc{red-bin}}& \textbf{\textsc{red-mul-5k}} \\
						\hline
						\hline
						\multicolumn{7}{c}{\textbf{Explicit inter-graph similarity based methods}}\\
						\hline
						DDGK & \underline{91.6 $\pm$ 6.8} & 63.1 $\pm$ 6.6 & $-$ &$-$ & $-$ &$-$\\
						GCKN-walk &  \textbf{92.8 $\pm$ 6.1}&\underline{65.9 $\pm$ 2.0}&\textbf{75.9 $\pm$ 3.7}&\textbf{53.4 $\pm$ 4.7}& $-$&$-$\\
						UGraphEmb & - & \textbf{72.5} & - & \underline{50.1}& $-$ &$-$\\
						\hline
						\multicolumn{7}{c}{\textbf{Non-Explicit inter-graph similarity based methods}}\\
						\multicolumn{7}{c}{Skip-gram based methods}\\
						\hline
						
						node2vec   & 72.6 $\pm$ 10.2  & 58.6 $\pm$ 8.0  & $-$  & $-$  & $-$ &-\\
						sub2vec  & 61.1 $\pm$ 15.8 & 60.0 $\pm$ 6.4 & 55.3 $\pm$ 1.5 & 36.7 $\pm$ 0.8 & 71.5 $\pm$ 0.4&36.7 $\pm$ 0.4\\
						graph2vec  & 83.2 $\pm$ 9.6 & 60.2 $\pm$ 6.9 &  71.1 $\pm$ 0.5 & 50.4 $\pm$ 0.9  & 75.8 $\pm$ 1.0&47.9 $\pm$ 0.3\\
						
						\hline
						\multicolumn{7}{c}{Contrastive Learning based methods}\\
						\hline
						
						 InfoGraph  & 89.0 $\pm$ 1.1 & 61.7 $\pm$ 1.4 & 73.0 $\pm$ 0.9 & 49.7 $\pm$ 0.5  & 82.5 $\pm$ 1.4&53.5 $\pm$ 1.0\\
						 CMV  &89.7 $\pm$ 1.1  &  62.5 $\pm$ 1.7 & \underline{74.2 $\pm$ 0.7} & \underline{51.2 $\pm$ 0.5} & 84.5 $\pm$ 0.6&$-$\\
						 GCC  & $-$  & $-$ & 72.0 & 49.4 & \underline{89.8}&53.7\\
						 GraphCL & 86.8 $\pm$ 1.3& $-$ & 71.1 $\pm$ 0.4 &$-$& 89.5 $\pm$ 0.4 & \textbf{56.0 $\pm$ 0.3}\\
						 
						 \hline
						\multicolumn{7}{c}{GVAE based methods}\\
						\hline
						 
						 GVAE(baseline) &87.7 $\pm$ 0.7&61.2$\pm$ 1.8 &70.7 $\pm$ 0.7&49.3 $\pm$ 0.4  &87.1 $\pm$ 0.1&52.8 $\pm$ 0.2\\
						 
						 \hline
						\multicolumn{7}{c}{\textbf{deepGCFX (Ours)} - $\alpha$ value for best results is in the brackets}\\
						\hline
						 
						 %\textbf{GCPLearner}(ours) &\textbf{90.7 $\pm$ 0.6}& \textbf{67.9 $\pm$ 0.8}& \textbf{74.7 $\pm$ 0.6}& \textbf{52.1 $\pm$ 0.4}&\textbf{90.9 $\pm$ 0.3}& \textbf{54.9 $\pm$ 0.1}\\
						 deepGCFX&\underline{89.8 $\pm$ 1.1}&\underline{66.5 $\pm$ 1.0}&72.9 $\pm$ 0.4&51.1$\pm$0.5&89.7$\pm$0.4&54.1$\pm$0.2\\
						 
						 %$\text{fused}(\mathbf{z}_\mathcal{C},\mathbf{z}_\mathcal{NC})$&&&&&&\\
						deepGCFX++&\textbf{92.2 $\pm$0.9(0.7)}&\textbf{69.6$\pm$ 1.4(0.85)}&\textbf{74.4$\pm$ 0.2 (0.95)} &\textbf{52.7$\pm$ 0.4(0.85)}&\textbf{90.9$\pm$ 0.3(0.9)}& \underline{55.1$\pm$ 0.2(0.85)}\\
						
					\end{tabular}
			\end{small}
		\end{center}
	\end{table*}
	
	\begin{table*}[h!]
  \caption{Mean node classification accuracy for supervised and unsupervised models for assortative and disassortative graphs. Results in \textbf{bold} indicate best supervised and unsupervised accuracy for each dataset and \underline{underlined} is the second best for unsupervised. We strictly follow the evaluation setup and datasets of Geom-GCN\cite{DBLP:conf/iclr/PeiWCLY20}. For unsupervised methods we used linear evaluation protocol.}
  \label{table:node_eval}
  \begin{center}
    \begin{small}
  
  \begin{tabular}{lccc  ccc}
    \hline
    & \multicolumn{3}{c}{\textbf{\textsc{assortative}}} & \multicolumn{3}{c}{\textbf{\textsc{disassortative}}} \\
    \hline
    \textbf{\textsc{datasets}} & \textbf{\textsc{cora}} &\textbf{\textsc{citeseer}} &\textbf{\textsc{pubmed}} &\textbf{\textsc{chameleon}} &\textbf{\textsc{squirrel}} &\textbf{\textsc{actor}}  \\
    \hline
	\multicolumn{6}{c}{Supervised Reference}\\
	\hline
	GCN &\textbf{85.77}&73.68&88.13&28.18&23.96& 26.86  \\
    Geom-GCN &85.27&\textbf{77.99}&\textbf{90.05}&\textbf{60.90}&\textbf{38.14}& \textbf{31.63}  \\
    \hline
	\multicolumn{6}{c}{Unsupervised baselines}\\
	\hline
    DGI &\textbf{82.16 $\pm$1.2}&\textbf{67.01$\pm$ 1.3}&\textbf{81.34$\pm$0.6} &\underline{59.45$\pm$2.4}&\underline{36.33$\pm$1.2}& \underline{27.09$\pm$ 1.2} \\
    GVAE &78.22 $\pm$1.4&63.9 $\pm$1.6&77.5$\pm$ 0.7&56.88$\pm$2.9&33.05$\pm$1.6& 25.12 $\pm$1.4  \\
    \hline
	\multicolumn{6}{c}{\textbf{deepGCFX - Ours}}\\
	\hline
    $\mathbf{z}_c$&30.33$\pm$ 1.2&20.75$\pm$ 1.1&39.82$\pm$ 0.5&19.30$\pm$2.7&
    19.23$\pm$0.8& 10.5$\pm$1.2 \\
    $\mathbf{Z}_l$ &81.26 $\pm$1.2&65.51 $\pm$1.4&79.85$\pm$ 0.7&57.67$\pm$3.1&    35.64$\pm$1.7&26.88$\pm$ 1.0\\
    %$\mathcal{D}(\mathbf{z}_\mathcal{NC},\mathbf{z}_\mathcal{C})$ &81.81 $\pm$1.14&66.78 $\pm$2.69&80.36$\pm$ 0.78&58.31$\pm$1.69& 36.78$\pm$1.31 & \\
    deepGCFX++ &\underline{81.96 $\pm$1.7}(0.15)&\underline{66.71$\pm$1.6}(0.1)&\underline{80.3$\pm$ 0.7}(0.2)& \textbf{61.05$\pm$2.4}(0.35)& \textbf{39.20$\pm$1.4}(0.4) & \textbf{28.80$\pm$ 1.4}(0.4)   \\
   
  \end{tabular}
  
  \end{small}
		\end{center}
\end{table*}

\section{Evaluation}

\subsection{Effectiveness analysis of deepGCFX for graph-wise common latent factor extraction}

\begin{figure}[h]
\centering
\subfloat[Correlation between $\mathbf{z}_c$ and $\mathbf{Z}_l$] {
  \includegraphics[width=42mm]{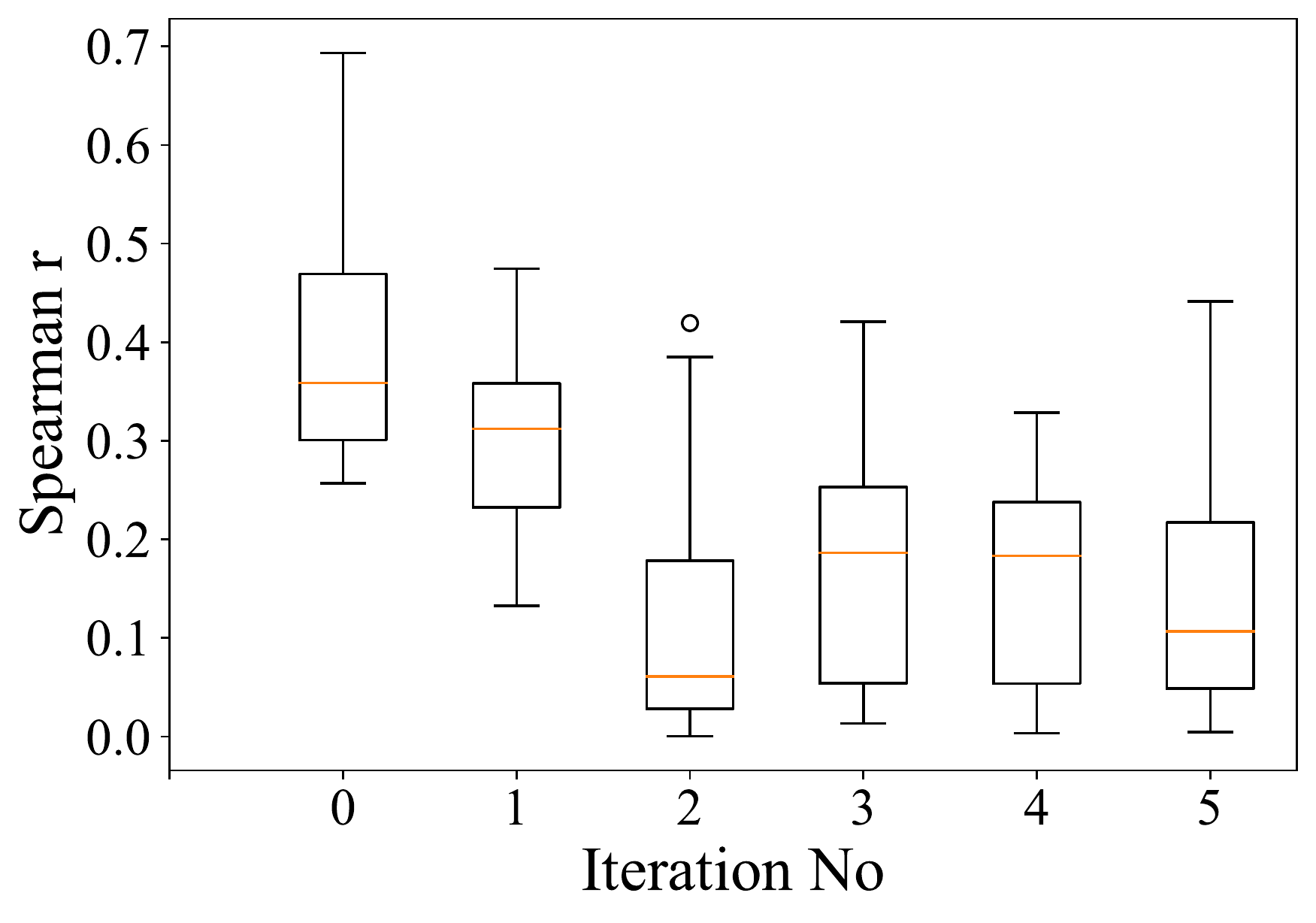}
}
\subfloat[Correlation between $\mathbf{z}_c$ and $\mathbf{Z}_c$] {
  \includegraphics[width=42mm]{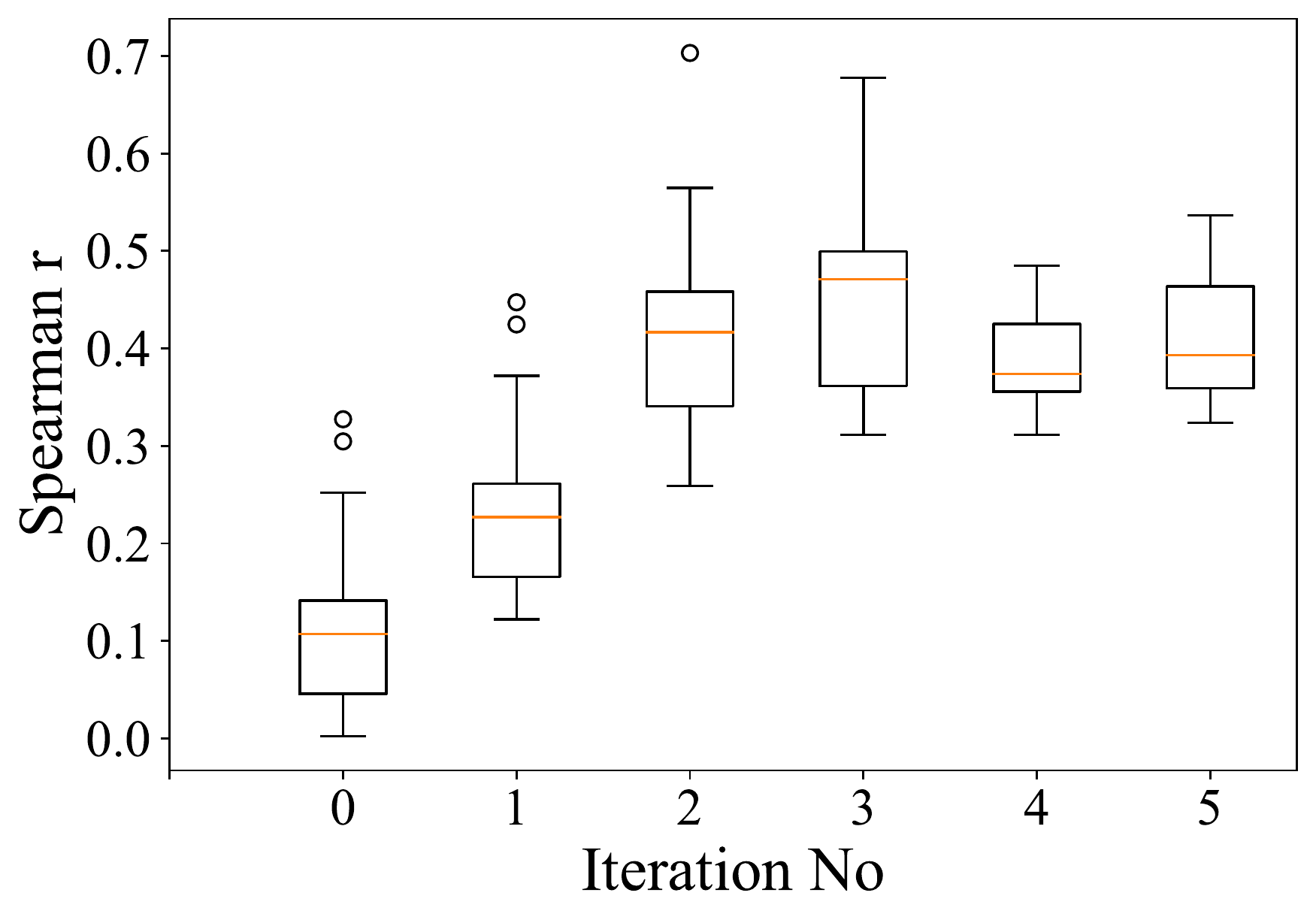}
}
\hspace{0mm}
\subfloat[Correlation between $\mathbf{z}_c$ and output from $\mathcal{D}_{agg}$] {
  \includegraphics[width=42mm]{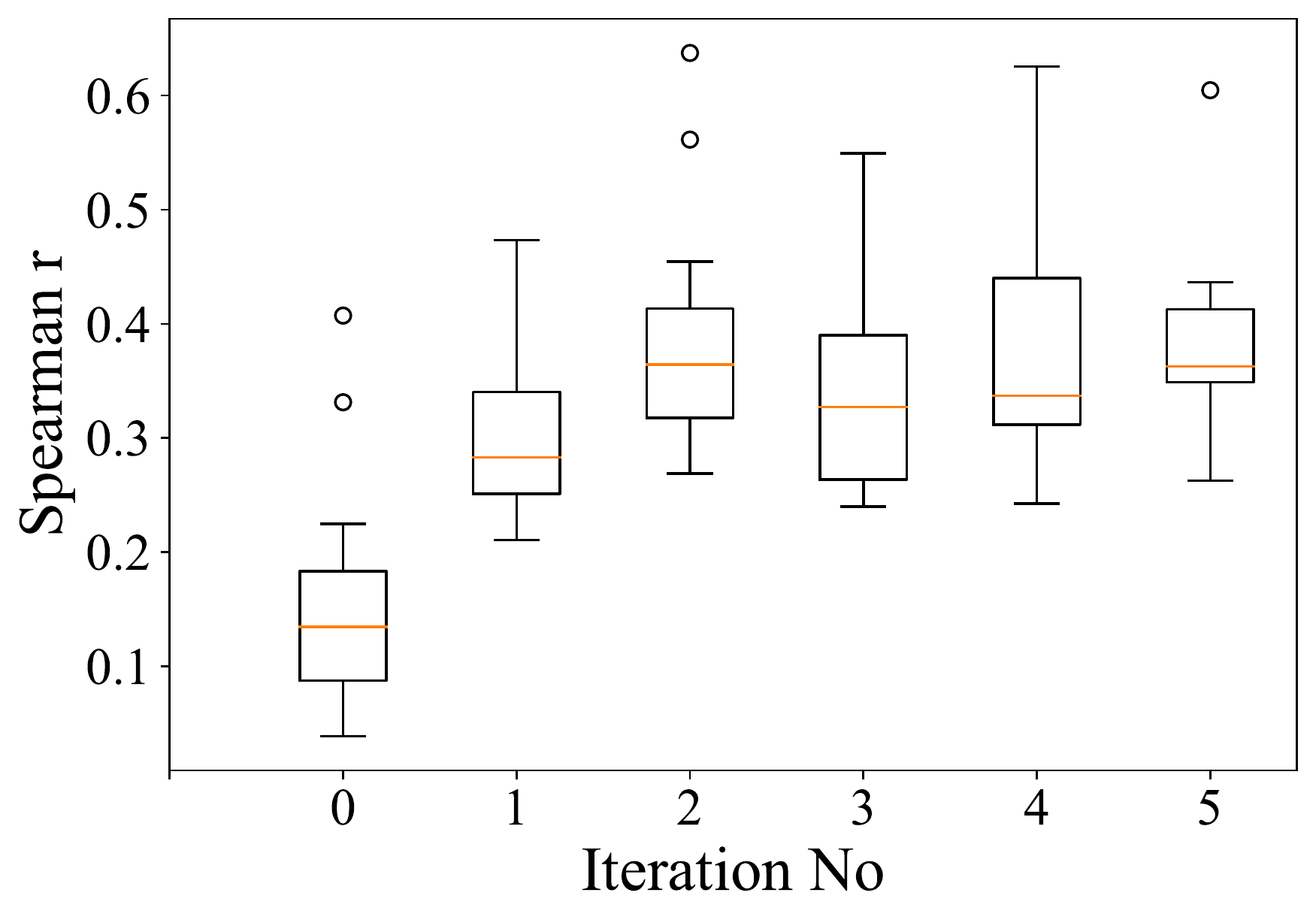}
}

\caption{Variation of Spearman R absolute correlation of extracted common latent factors $\mathbf{z}_c$ with local ($\mathbf{Z}_l$), patch-specific common $\mathbf{Z}_c$ and the output from $\mathcal{D}_{agg}$ against number ACCUM iterations.}
\label{fig:corre}
\end{figure}

To evaluate the effectiveness of deepGCFX in extracting graph-wise common latent factors, we analyze how the correlation of extracted common latent $\mathbf{z}_c$ varies with local latent factors $\mathbf{Z}_l$, patch-specific common latent factors $\mathbf{Z}_c$ and the output from the aggregation decoder $\mathcal{D}_{agg}$. Following graph recovery methods such as UDR\cite{DBLP:conf/iclr/DuanMSWBLH20}, we use Spearman's correlation for this. Fig. \ref{fig:corre} shows how the correlation changes with the number of iterations for a sample graph from the MUTAG dataset. We start with the accumulation iteration 0 where the common and local latent filtering is done at random to show how well the proposed ACCUM functions against random filtering. 

\begin{figure}[h]
    \centering
    \subfloat[\centering Correlation between common and local factors from deepGCFX (left) vs GVAE (right) ]{{\includegraphics[width=.7\linewidth]{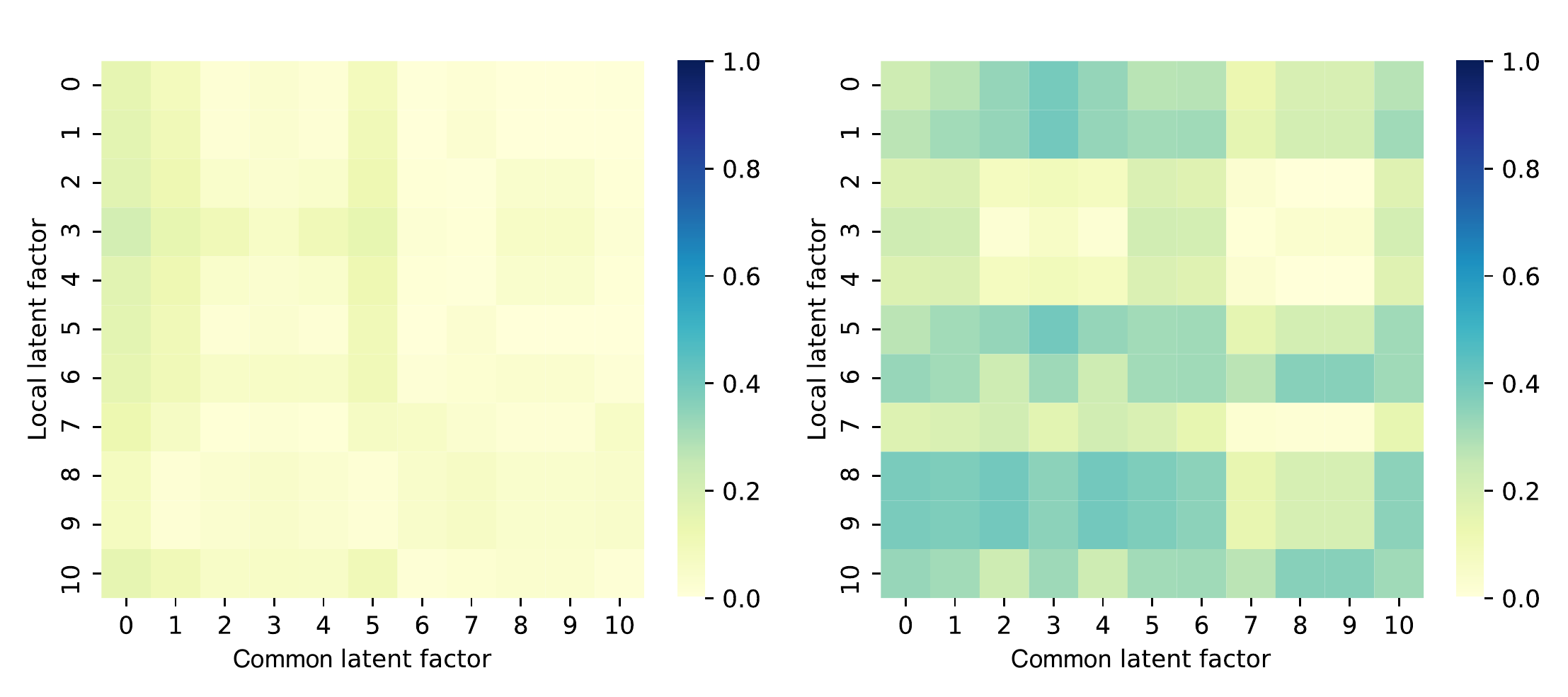} }}%
    \qquad
    \subfloat[\centering Patch-wise MAPD for Common(left) and Local(right) factors ]{{\includegraphics[width=.7\linewidth]{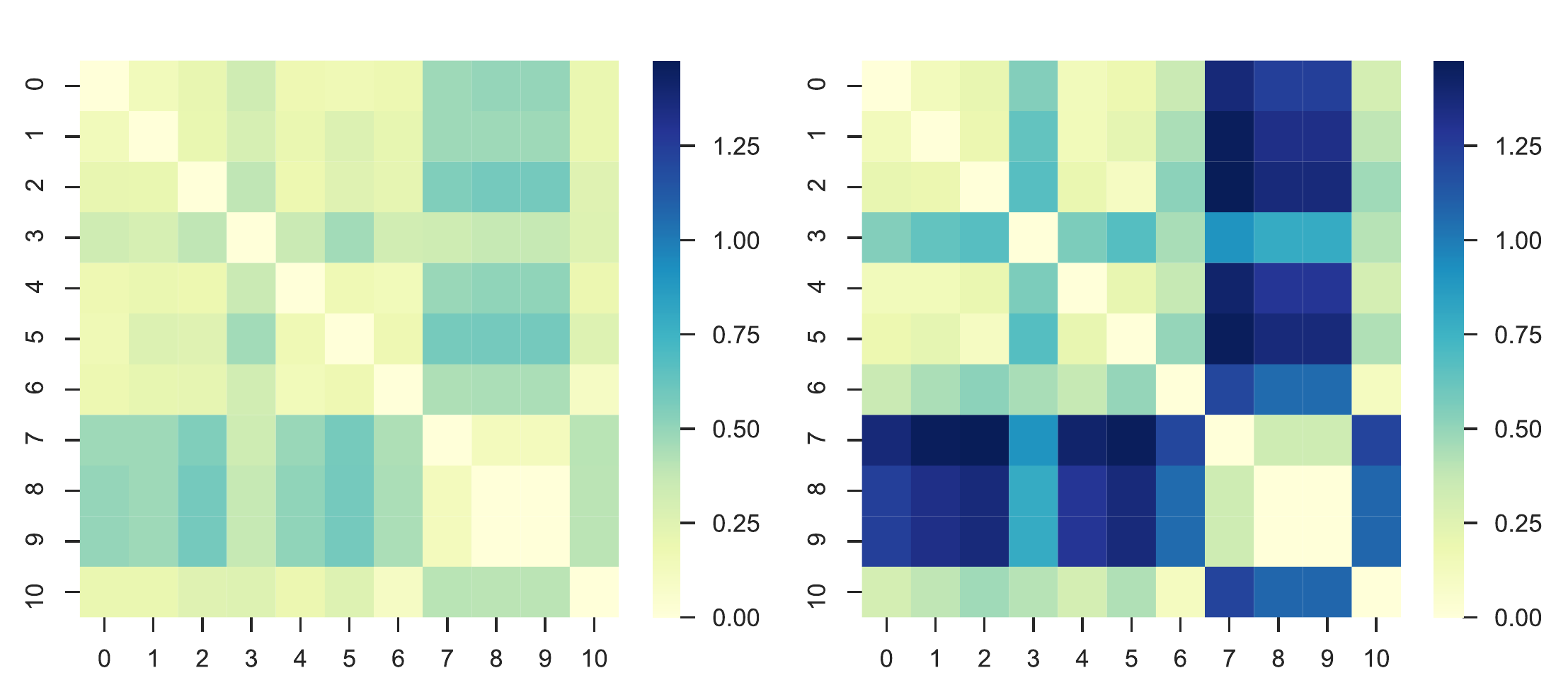} }}%
    \caption{Analysis on deepGCFX: (a) Filtering ability of learnt Common and Local latent factors by deepGCFX via absolute values of correlation compared with GVAE, which does not perform common-local filtering. (b) Inter-patch MAPD among Common and Local latent factors. Lower MAPD for $\mathbf{z}_c$ indicates the common factor representations filtered by deepGCFX is indeed shared across patches of the entire graph, unlike $\mathbf{Z}_l$ which are specific to certain patches.}
    \label{fig:pariwise_diff}
\end{figure}

We can observe that with the increase of ACCUM iterations, the correlation between local factors $\mathbf{Z}_l$ and common factors $\mathbf{z}_c$ decreases (Fig. \ref{fig:corre}(a)) showing our proposed ACCUM method's effectiveness in feature differentiation  by being able to filter different information as common and local latent factors. Fig. \ref{fig:corre}(b) shows the positive relationship common latent has with its patch-specific factors as expected. Finally, in Fig. \ref{fig:corre}(c) the correlation of $\mathbf{z}_c$ with the output of $\mathcal{D}_{agg}$ increases showing that $\mathcal{D}_{agg}$ considers common graph-level factors without overemphasizing on local proximity, hence ensuring commonality of $\mathbf{z}_c$. This shows the effectiveness of iterative ACCUM in extracting common factors.

In Fig. \ref{fig:pariwise_diff}(a) we compare deepGCFX's filtering ability of $\mathbf{z}_c$ from $\mathbf{z}_l$ against GVAE which does not have that ability. To obtain two latent representations from GVAE, we divide the learnt latent into halves. We obtain Spearman R based correlation matrix and by analysing them we can conclude that, deepGCFX indeed can differentiate features, as the correlation between $\mathbf{z}_c$ from $\mathbf{z}_l$ is very low compared to GVAE. To verify the factors extracted by our query-based reasoning ACCUM indeed are common across the graph, we use the experiment in Fig. \ref{fig:pariwise_diff}(b). We use Mean Absolute Pairwise Difference (MAPD) measure used by \citet{DBLP:conf/iclr/HigginsMPBGBML17} to compare the inter-patch similarity of common and local latent separately and we can observe that inter-patch common latent factor similarity is very high compared to inter-patch local factors demonstrating that the latent factors extracted by our ACCUM method are indeed common across the graph.

\subsection{Impact of learnt graph-wise common factors $\mathbf{z}_c$ and local factors $\mathbf{z}_l$ on Downstream task performance.}

To evaluate the discriminative ability of extracted common latent factors $\mathbf{z}_c$ on downstream tasks, we select graph classification. We report results for deepGCFX when only $\mathbf{z}_c$ is used as the graph embedding. deepGCFX++ combines both common and local factors with a gating mechanism as $\alpha \mathbf{z}_c+(1-\alpha)\sum_{j=1}^{|{V}|} \mathbf{z}_l(j)$, where $\alpha$ denotes the contribution from graph-wise common factors. We compare deepGCFX with existing state-of-the-art methods and report results in Table \ref{table:graph_eval}. Compared to existing work on skip-gram and contrastive learning, deepGCFX achieves comparable or better results for four datasets and our deepGCFX++(when the biggest contribution comes from common latent factors) achieves the state-of-the-art results for 5 datasets and is very competitive with GraphCL\cite{DBLP:conf/nips/YouCSCWS20} which uses data augmentations for REDDI-MULTI-5K showing the effectiveness of GCFX over contrastive learning, whose model performance relies on the selection of negative samples.  Explicit inter-graph similarity based methods have not reported results for larger datasets like REDDIT due to the high computational cost of pair-wise comparisons. For other datasets, we achieve competitive results with them by just using a single graph sample for embedding learning. These results show the effectiveness of utilizing graph-wise common factors as graph embeddings.

\subsection{Impact of $\mathbf{z}_c$ for node-level tasks}

To analyze how common latent factors affect node-level tasks, we select node classification task on both assortative (neighbour nodes of a graph have same class label) and disassortative (neighbour nodes have different class labels) graph. Since $\mathbf{z}_c$ is common for all the nodes in the graph despite the inter-node distance, we want to analyze whether it is beneficial in improving long-range node dependencies. Non-local aggregating \cite{DBLP:conf/iclr/PeiWCLY20,liu2020non} for graph learning has drawn attention in supervised graph learning research due to GNN's inability in long-distance information propagation. Geom-GCN \cite{DBLP:conf/iclr/PeiWCLY20} proposed a benchmark to evaluate non-local aggregation for both assortative and disassortative graphs.

To the best of our knowledge, existing unsupervised graph representation learning methods have not used this benchmark for evaluation so far, hence we select two supervised (GCN\cite{DBLP:conf/iclr/KipfW17}, Geom-GCN\cite{DBLP:conf/iclr/PeiWCLY20}) for the reference and two unsupervised (Deep Graph Infomax (DGI)\cite{DBLP:conf/iclr/VelickovicFHLBH19}, GVAE\cite{DBLP:journals/corr/KipfW16a}) methods as our baselines. Table \ref{table:node_eval} reports results for all models and, compared to local only ($\mathbf{Z}_l$) of deepGCFX, deepGCFX++ which incorporates common latent factors with local factors achieves higher results demonstrating the effectiveness of common latent factors for node-level tasks. More interestingly, deepGCFX++ has achieved best results for disassortative graphs highlighting extracted common latent factors' ability in capturing long-distance node dependencies.

\section{Conclusion}
We introduced a graph-wise common latent factor extraction based principle for unsupervised graph representation learning. Based on real-world graphs,  we identified  common factors that are used along with node-specific local factors for graph generation and we hypothesized extracting common factors could be highly beneficial for discriminative graph representations. We proposed GCFX principle and deepGCFX model to address feature differentiation and proximity overemphasis limitations of GVAE to enable graph-wise common latent factor extraction. With extensive experiments, we demonstrated the effectiveness of deepGCFX and its outstanding performance over state-of-the-art methods with only using current sample for embedding learning.

\section*{Acknowledgment}
This work was supported by SUTD project PIE-SGP-AI-2018-01. This research was also supported by the National Research Foundation Singapore under its AI Singapore Programme [Award Number:AISG-100E2018-005]. The authors are grateful for the discussion with Lu Wei.

\bibliography{aaai22}

\begin{thebibliography}{45}
\providecommand{\natexlab}[1]{#1}

\bibitem[{Adhikari et~al.(2018)Adhikari, Zhang, Ramakrishnan, and
  Prakash}]{DBLP:conf/pakdd/AdhikariZRP18}
Adhikari, B.; Zhang, Y.; Ramakrishnan, N.; and Prakash, B.~A. 2018.
\newblock Sub2Vec: Feature Learning for Subgraphs.
\newblock In \emph{Advances in Knowledge Discovery and Data Mining - 22nd
  Pacific-Asia Conference, {PAKDD} 2018, Melbourne, VIC, Australia, June 3-6,
  2018, Proceedings, Part {II}}, 170--182.

\bibitem[{Al{-}Rfou, Perozzi, and Zelle(2019)}]{DBLP:conf/www/Al-RfouPZ19}
Al{-}Rfou, R.; Perozzi, B.; and Zelle, D. 2019.
\newblock {DDGK:} Learning Graph Representations for Deep Divergence Graph
  Kernels.
\newblock In Liu, L.; White, R.~W.; Mantrach, A.; Silvestri, F.; McAuley,
  J.~J.; Baeza{-}Yates, R.; and Zia, L., eds., \emph{The World Wide Web
  Conference, {WWW} 2019, San Francisco, CA, USA, May 13-17, 2019}, 37--48.
  {ACM}.

\bibitem[{Bai et~al.(2019)Bai, Ding, Qiao, Marinovic, Gu, Chen, Sun, and
  Wang}]{DBLP:conf/ijcai/BaiDQMG0SW19}
Bai, Y.; Ding, H.; Qiao, Y.; Marinovic, A.; Gu, K.; Chen, T.; Sun, Y.; and
  Wang, W. 2019.
\newblock Unsupervised Inductive Graph-Level Representation Learning via
  Graph-Graph Proximity.
\newblock In \emph{Proceedings of the Twenty-Eighth International Joint
  Conference on Artificial Intelligence, {IJCAI} 2019, Macao, China, August
  10-16, 2019}, 1988--1994.

\bibitem[{Baldi and Hornik(1989)}]{Baldi1989NeuralNA}
Baldi, P.; and Hornik, K. 1989.
\newblock Neural networks and principal component analysis: Learning from
  examples without local minima.
\newblock \emph{Neural Networks}, 2: 53--58.

\bibitem[{Bhatia et~al.(2016)Bhatia, Dahiya, Jain, Mittal, Prabhu, and
  Varma}]{Bhatia16}
Bhatia, K.; Dahiya, K.; Jain, H.; Mittal, A.; Prabhu, Y.; and Varma, M. 2016.
\newblock The extreme classification repository: Multi-label datasets and code.

\bibitem[{{Borgwardt} and {Kriegel}(2005)}]{1565664}
{Borgwardt}, K.~M.; and {Kriegel}, H.~P. 2005.
\newblock Shortest-path kernels on graphs.
\newblock In \emph{Fifth IEEE International Conference on Data Mining
  (ICDM'05)}, 8 pp.--.

\bibitem[{Brown et~al.(2019)Brown, Fiscato, Segler, and
  Vaucher}]{doi:10.1021/acs.jcim.8b00839}
Brown, N.; Fiscato, M.; Segler, M.~H.; and Vaucher, A.~C. 2019.
\newblock GuacaMol: Benchmarking Models for de Novo Molecular Design.
\newblock \emph{Journal of Chemical Information and Modeling}, 59(3):
  1096--1108.

\bibitem[{Chen, Jacob, and Mairal(2020)}]{DBLP:journals/corr/abs-2003-05189}
Chen, D.; Jacob, L.; and Mairal, J. 2020.
\newblock Convolutional Kernel Networks for Graph-Structured Data.
\newblock \emph{CoRR}, abs/2003.05189.

\bibitem[{Cooray, Cheung, and Lu(2020)}]{Cooray_2020_CVPR}
Cooray, T.; Cheung, N.-M.; and Lu, W. 2020.
\newblock Attention-Based Context Aware Reasoning for Situation Recognition.
\newblock In \emph{The IEEE/CVF Conference on Computer Vision and Pattern
  Recognition (CVPR)}.

\bibitem[{Duan et~al.(2020)Duan, Matthey, Saraiva, Watters, Burgess, Lerchner,
  and Higgins}]{DBLP:conf/iclr/DuanMSWBLH20}
Duan, S.; Matthey, L.; Saraiva, A.; Watters, N.; Burgess, C.; Lerchner, A.; and
  Higgins, I. 2020.
\newblock Unsupervised Model Selection for Variational Disentangled
  Representation Learning.
\newblock In \emph{8th International Conference on Learning Representations,
  {ICLR} 2020, Addis Ababa, Ethiopia, April 26-30, 2020}.

\bibitem[{Duvenaud et~al.(2015)Duvenaud, Maclaurin, Iparraguirre, Bombarell,
  Hirzel, Aspuru-Guzik, and Adams}]{NIPS2015_5954}
Duvenaud, D.~K.; Maclaurin, D.; Iparraguirre, J.; Bombarell, R.; Hirzel, T.;
  Aspuru-Guzik, A.; and Adams, R.~P. 2015.
\newblock Convolutional Networks on Graphs for Learning Molecular Fingerprints.
\newblock In Cortes, C.; Lawrence, N.~D.; Lee, D.~D.; Sugiyama, M.; and
  Garnett, R., eds., \emph{Advances in Neural Information Processing Systems
  28}, 2224--2232. Curran Associates, Inc.

\bibitem[{G{\"{a}}rtner, Flach, and Wrobel(2003)}]{DBLP:conf/colt/GartnerFW03}
G{\"{a}}rtner, T.; Flach, P.~A.; and Wrobel, S. 2003.
\newblock On Graph Kernels: Hardness Results and Efficient Alternatives.
\newblock In \emph{Computational Learning Theory and Kernel Machines, 16th
  Annual Conference on Computational Learning Theory and 7th Kernel Workshop,
  COLT/Kernel 2003, Washington, DC, USA, August 24-27, 2003, Proceedings},
  129--143.

\bibitem[{Grill et~al.(2020)Grill, Strub, Altch{\'{e}}, Tallec, Richemond,
  Buchatskaya, Doersch, Pires, Guo, Azar, Piot, Kavukcuoglu, Munos, and
  Valko}]{DBLP:conf/nips/GrillSATRBDPGAP20}
Grill, J.; Strub, F.; Altch{\'{e}}, F.; Tallec, C.; Richemond, P.~H.;
  Buchatskaya, E.; Doersch, C.; Pires, B.~{\'{A}}.; Guo, Z.; Azar, M.~G.; Piot,
  B.; Kavukcuoglu, K.; Munos, R.; and Valko, M. 2020.
\newblock Bootstrap Your Own Latent - {A} New Approach to Self-Supervised
  Learning.
\newblock In Larochelle, H.; Ranzato, M.; Hadsell, R.; Balcan, M.; and Lin, H.,
  eds., \emph{Advances in Neural Information Processing Systems 33: Annual
  Conference on Neural Information Processing Systems 2020, NeurIPS 2020,
  December 6-12, 2020, virtual}.

\bibitem[{Grover and Leskovec(2016)}]{DBLP:conf/kdd/GroverL16}
Grover, A.; and Leskovec, J. 2016.
\newblock node2vec: Scalable Feature Learning for Networks.
\newblock In \emph{Proceedings of the 22nd {ACM} {SIGKDD} International
  Conference on Knowledge Discovery and Data Mining, San Francisco, CA, USA,
  August 13-17, 2016}, 855--864.

\bibitem[{Hassani and Khasahmadi(2020)}]{icml2020_1971}
Hassani, K.; and Khasahmadi, A.~H. 2020.
\newblock Contrastive Multi-View Representation Learning on Graphs.
\newblock In \emph{Proceedings of International Conference on Machine
  Learning}, 3451--3461.

\bibitem[{Higgins et~al.(2017)Higgins, {\ "{\ i}}~c Matthey, Pal, Burgess,
  Glorot, Botvinick, Mohamed, and Lerchner}]{DBLP:conf/iclr/HigginsMPBGBML17}
Higgins, I.; {\ "{\ i}}~c Matthey, L.; Pal, A.; Burgess, C.; Glorot, X.;
  Botvinick, M.; Mohamed, S.; and Lerchner, A. 2017.
\newblock beta-VAE: Learning Basic Visual Concepts with a Constrained
  Variational Framework.
\newblock In \emph{5th International Conference on Learning Representations,
  {ICLR} 2017, Toulon, France, April 24-26, 2017, Conference Track
  Proceedings}.

\bibitem[{Hinton and Zemel(1993)}]{DBLP:conf/nips/HintonZ93}
Hinton, G.~E.; and Zemel, R.~S. 1993.
\newblock Autoencoders, Minimum Description Length and Helmholtz Free Energy.
\newblock In \emph{Advances in Neural Information Processing Systems 6, [7th
  {NIPS} Conference, Denver, Colorado, USA, 1993]}, 3--10.

\bibitem[{Kipf and Welling(2016)}]{DBLP:journals/corr/KipfW16a}
Kipf, T.~N.; and Welling, M. 2016.
\newblock Variational Graph Auto-Encoders.
\newblock \emph{CoRR}, abs/1611.07308.

\bibitem[{Kipf and Welling(2017)}]{DBLP:conf/iclr/KipfW17}
Kipf, T.~N.; and Welling, M. 2017.
\newblock Semi-Supervised Classification with Graph Convolutional Networks.
\newblock In \emph{5th International Conference on Learning Representations,
  {ICLR} 2017, Toulon, France, April 24-26, 2017, Conference Track
  Proceedings}.

\bibitem[{Krishna et~al.(2016)Krishna, Zhu, Groth, Johnson, Hata, Kravitz,
  Chen, Kalantidis, Li, Shamma, Bernstein, and Fei-Fei}]{krishnavisualgenome}
Krishna, R.; Zhu, Y.; Groth, O.; Johnson, J.; Hata, K.; Kravitz, J.; Chen, S.;
  Kalantidis, Y.; Li, L.-J.; Shamma, D.~A.; Bernstein, M.; and Fei-Fei, L.
  2016.
\newblock Visual Genome: Connecting Language and Vision Using Crowdsourced
  Dense Image Annotations.

\bibitem[{Krogan et~al.(2006)Krogan, Cagney, Yu, Zhong, Guo, Ignatchenko, Li,
  Pu, Datta, Tikuisis, Punna, Peregrín-Alvarez, Shales, Zhang, Davey,
  Robinson, Paccanaro, Bray, Sheung, and Greenblatt}]{ppifirst}
Krogan, N.; Cagney, G.; Yu, H.; Zhong, G.; Guo, X.; Ignatchenko, A.; Li, J.;
  Pu, S.; Datta, N.; Tikuisis, A.; Punna, T.; Peregrín-Alvarez, J.; Shales,
  M.; Zhang, X.; Davey, M.; Robinson, M.; Paccanaro, A.; Bray, J.; Sheung, A.;
  and Greenblatt, J. 2006.
\newblock Global landscape of protein complexes in the yeast Saccharomyces
  cerevisiae.
\newblock \emph{Nature}, 440: 637--43.

\bibitem[{Kuhn and Beratan(1996)}]{kuhn1996inverse}
Kuhn, C.; and Beratan, D.~N. 1996.
\newblock Inverse strategies for molecular design.
\newblock \emph{The Journal of Physical Chemistry}, 100(25): 10595--10599.

\bibitem[{Linsker(1988)}]{DBLP:journals/computer/Linsker88}
Linsker, R. 1988.
\newblock Self-Organization in a Perceptual Network.
\newblock \emph{computer}, 21(3): 105--117.

\bibitem[{Liu, Wang, and Ji(2020)}]{liu2020non}
Liu, M.; Wang, Z.; and Ji, S. 2020.
\newblock Non-local graph neural networks.
\newblock \emph{arXiv preprint arXiv:2005.14612}.

\bibitem[{Liu and Cheung(2021)}]{DBLP:journals/sigpro/LiuC21}
Liu, R.; and Cheung, N. 2021.
\newblock Joint estimation of low-rank components and connectivity graph in
  high-dimensional graph signals: Application to brain imaging.
\newblock \emph{Signal Process.}, 182: 107931.

\bibitem[{Narayanan et~al.(2017)Narayanan, Chandramohan, Venkatesan, Chen, Liu,
  and Jaiswal}]{DBLP:journals/corr/NarayananCVCLJ17}
Narayanan, A.; Chandramohan, M.; Venkatesan, R.; Chen, L.; Liu, Y.; and
  Jaiswal, S. 2017.
\newblock graph2vec: Learning Distributed Representations of Graphs.
\newblock \emph{CoRR}, abs/1707.05005.

\bibitem[{Newman and Girvan(2004)}]{PhysRevE.69.026113}
Newman, M. E.~J.; and Girvan, M. 2004.
\newblock Finding and evaluating community structure in networks.
\newblock \emph{Phys. Rev. E}, 69: 026113.

\bibitem[{Pan et~al.(2018)Pan, Hu, Long, Jiang, Yao, and
  Zhang}]{DBLP:conf/ijcai/PanHLJYZ18}
Pan, S.; Hu, R.; Long, G.; Jiang, J.; Yao, L.; and Zhang, C. 2018.
\newblock Adversarially Regularized Graph Autoencoder for Graph Embedding.
\newblock In \emph{Proceedings of the Twenty-Seventh International Joint
  Conference on Artificial Intelligence, {IJCAI} 2018, July 13-19, 2018,
  Stockholm, Sweden}, 2609--2615.

\bibitem[{Pang and Cheung(2017)}]{7814302}
Pang, J.; and Cheung, G. 2017.
\newblock Graph Laplacian Regularization for Image Denoising: Analysis in the
  Continuous Domain.
\newblock \emph{IEEE Transactions on Image Processing}, 26(4): 1770--1785.

\bibitem[{Park et~al.(2019)Park, Lee, Chang, Lee, and
  Choi}]{DBLP:conf/iccv/ParkLCLC19}
Park, J.; Lee, M.; Chang, H.~J.; Lee, K.; and Choi, J.~Y. 2019.
\newblock Symmetric Graph Convolutional Autoencoder for Unsupervised Graph
  Representation Learning.
\newblock In \emph{2019 {IEEE/CVF} International Conference on Computer Vision,
  {ICCV} 2019, Seoul, Korea (South), October 27 - November 2, 2019},
  6518--6527.

\bibitem[{Pei et~al.(2020)Pei, Wei, Chang, Lei, and
  Yang}]{DBLP:conf/iclr/PeiWCLY20}
Pei, H.; Wei, B.; Chang, K.~C.; Lei, Y.; and Yang, B. 2020.
\newblock Geom-GCN: Geometric Graph Convolutional Networks.
\newblock In \emph{8th International Conference on Learning Representations,
  {ICLR} 2020, Addis Ababa, Ethiopia, April 26-30, 2020}. OpenReview.net.

\bibitem[{Qiu et~al.(2020)Qiu, Chen, Dong, Zhang, Yang, Ding, Wang, and
  Tang}]{qiu2020gcc}
Qiu, J.; Chen, Q.; Dong, Y.; Zhang, J.; Yang, H.; Ding, M.; Wang, K.; and Tang,
  J. 2020.
\newblock GCC: Graph Contrastive Coding for Graph Neural Network Pre-Training.
\newblock \emph{arXiv preprint arXiv:2006.09963}.

\bibitem[{Sanchez-Lengeling and Aspuru-Guzik(2018)}]{Sanchez-Lengeling360}
Sanchez-Lengeling, B.; and Aspuru-Guzik, A. 2018.
\newblock Inverse molecular design using machine learning: Generative models
  for matter engineering.
\newblock \emph{Science}, 361(6400): 360--365.

\bibitem[{Schneider(2013)}]{schneider2013novo}
Schneider, G. 2013.
\newblock \emph{De novo molecular design}.
\newblock John Wiley \& Sons.

\bibitem[{Shervashidze et~al.(2009)Shervashidze, Vishwanathan, Petri, Mehlhorn,
  and Borgwardt}]{DBLP:journals/jmlr/ShervashidzeVPMB09}
Shervashidze, N.; Vishwanathan, S. V.~N.; Petri, T.; Mehlhorn, K.; and
  Borgwardt, K.~M. 2009.
\newblock Efficient graphlet kernels for large graph comparison.
\newblock In \emph{Proceedings of the Twelfth International Conference on
  Artificial Intelligence and Statistics, {AISTATS} 2009, Clearwater Beach,
  Florida, USA, April 16-18, 2009}, 488--495.

\bibitem[{Sun et~al.(2020)Sun, Hoffmann, Verma, and
  Tang}]{DBLP:conf/iclr/SunHV020}
Sun, F.; Hoffmann, J.; Verma, V.; and Tang, J. 2020.
\newblock InfoGraph: Unsupervised and Semi-supervised Graph-Level
  Representation Learning via Mutual Information Maximization.
\newblock In \emph{8th International Conference on Learning Representations,
  {ICLR} 2020, Addis Ababa, Ethiopia, April 26-30, 2020}.

\bibitem[{Tian, Krishnan, and Isola(2020)}]{DBLP:conf/eccv/TianKI20}
Tian, Y.; Krishnan, D.; and Isola, P. 2020.
\newblock Contrastive Multiview Coding.
\newblock In Vedaldi, A.; Bischof, H.; Brox, T.; and Frahm, J., eds.,
  \emph{Computer Vision - {ECCV} 2020 - 16th European Conference, Glasgow, UK,
  August 23-28, 2020, Proceedings, Part {XI}}, volume 12356 of \emph{Lecture
  Notes in Computer Science}, 776--794. Springer.

\bibitem[{Velickovic et~al.(2018)Velickovic, Cucurull, Casanova, Romero, {\
  `{o}}, and Bengio}]{DBLP:conf/iclr/VelickovicCCRLB18}
Velickovic, P.; Cucurull, G.; Casanova, A.; Romero, A.; {\ `{o}}, P.~L.; and
  Bengio, Y. 2018.
\newblock Graph Attention Networks.
\newblock In \emph{6th International Conference on Learning Representations,
  {ICLR} 2018, Vancouver, BC, Canada, April 30 - May 3, 2018, Conference Track
  Proceedings}.

\bibitem[{Velickovic et~al.(2019)Velickovic, Fedus, Hamilton, Li{\`{o}},
  Bengio, and Hjelm}]{DBLP:conf/iclr/VelickovicFHLBH19}
Velickovic, P.; Fedus, W.; Hamilton, W.~L.; Li{\`{o}}, P.; Bengio, Y.; and
  Hjelm, R.~D. 2019.
\newblock Deep Graph Infomax.
\newblock In \emph{7th International Conference on Learning Representations,
  {ICLR} 2019, New Orleans, LA, USA, May 6-9, 2019}.

\bibitem[{Wieder et~al.(2020)Wieder, Kohlbacher, Kuenemann, Garon, Ducrot,
  Seidel, and Langer}]{WIEDER2020}
Wieder, O.; Kohlbacher, S.; Kuenemann, M.; Garon, A.; Ducrot, P.; Seidel, T.;
  and Langer, T. 2020.
\newblock A compact review of molecular property prediction with graph neural
  networks.
\newblock \emph{Drug Discovery Today: Technologies}.

\bibitem[{Xu et~al.(2019)Xu, Hu, Leskovec, and
  Jegelka}]{DBLP:conf/iclr/XuHLJ19}
Xu, K.; Hu, W.; Leskovec, J.; and Jegelka, S. 2019.
\newblock How Powerful are Graph Neural Networks?
\newblock In \emph{7th International Conference on Learning Representations,
  {ICLR} 2019, New Orleans, LA, USA, May 6-9, 2019}.

\bibitem[{Yanardag and Vishwanathan(2015)}]{DBLP:conf/kdd/YanardagV15}
Yanardag, P.; and Vishwanathan, S. V.~N. 2015.
\newblock Deep Graph Kernels.
\newblock In Cao, L.; Zhang, C.; Joachims, T.; Webb, G.~I.; Margineantu, D.~D.;
  and Williams, G., eds., \emph{Proceedings of the 21th {ACM} {SIGKDD}
  International Conference on Knowledge Discovery and Data Mining, Sydney, NSW,
  Australia, August 10-13, 2015}, 1365--1374. {ACM}.

\bibitem[{Yang et~al.(2019)Yang, Swanson, Jin, Coley, Eiden, Gao, Guzman-Perez,
  Hopper, Kelley, Mathea et~al.}]{yang2019analyzing}
Yang, K.; Swanson, K.; Jin, W.; Coley, C.; Eiden, P.; Gao, H.; Guzman-Perez,
  A.; Hopper, T.; Kelley, B.; Mathea, M.; et~al. 2019.
\newblock Analyzing learned molecular representations for property prediction.
\newblock \emph{Journal of chemical information and modeling}, 59(8):
  3370--3388.

\bibitem[{You et~al.(2020)You, Chen, Sui, Chen, Wang, and
  Shen}]{DBLP:conf/nips/YouCSCWS20}
You, Y.; Chen, T.; Sui, Y.; Chen, T.; Wang, Z.; and Shen, Y. 2020.
\newblock Graph Contrastive Learning with Augmentations.
\newblock In Larochelle, H.; Ranzato, M.; Hadsell, R.; Balcan, M.; and Lin, H.,
  eds., \emph{Advances in Neural Information Processing Systems 33: Annual
  Conference on Neural Information Processing Systems 2020, NeurIPS 2020,
  December 6-12, 2020, virtual}.

\bibitem[{Zunger(2018)}]{zunger2018inverse}
Zunger, A. 2018.
\newblock Inverse design in search of materials with target functionalities.
\newblock \emph{Nature Reviews Chemistry}, 2(4): 1--16.

\end{thebibliography}

\end{document}